\documentclass[10pt,twocolumn,letterpaper]{article}

\usepackage[pagenumbers]{cvpr} 

\usepackage{epsfig}
\usepackage{enumitem}
\usepackage{threeparttable}
\usepackage{color}

\usepackage{graphicx}
\usepackage{caption}
\usepackage{amsmath}
\usepackage{amssymb}
\usepackage{booktabs}
\usepackage{adjustbox} 
\usepackage{multirow}
\usepackage{diagbox}
\usepackage{float}
\usepackage{colortbl}
\usepackage{xcolor}
\usepackage{array}
\usepackage[pagebackref,breaklinks,colorlinks]{hyperref}

\usepackage[capitalize]{cleveref}
\crefname{section}{Sec.}{Secs.}
\Crefname{section}{Section}{Sections}
\Crefname{table}{Table}{Tables}
\crefname{table}{Tab.}{Tabs.}

\def\cvprPaperID{9105} 
\def\confName{CVPR}
\def\confYear{2023}

\begin{document}

\title{ UniOcc: Unifying Vision-Centric 3D Occupancy Prediction \\ with Geometric and Semantic Rendering}


\author{
Mingjie Pan\textsuperscript{\rm 1,2},
Li Liu\textsuperscript{\rm 1},
Jiaming Liu\textsuperscript{\rm 2},
Peixiang Huang\textsuperscript{\rm 1,2}, \\
Longlong Wang\textsuperscript{\rm 1},
Shaoqing Xu\textsuperscript{\rm 1},
Zhiyi Lai\textsuperscript{\rm 1},
Shanghang Zhang\textsuperscript{\rm 2}\thanks{Corresponding author},
Kuiyuan Yang\textsuperscript{\rm 1}, \\
\textsuperscript{\rm 1}Xiaomi Car, 
\textsuperscript{\rm 2}National Key Laboratory for Multimedia Information Processing, Peking University\\ 
}

\maketitle

\begin{abstract}
In this technical report, we present our solution, named UniOCC, for the Vision-Centric 3D occupancy prediction track in the nuScenes Open Dataset Challenge at CVPR 2023.
Existing methods for occupancy prediction primarily focus on optimizing projected features on 3D volume space using 3D occupancy labels. 
However, the generation process of these labels is complex and expensive (relying on 3D semantic annotations), and limited by voxel resolution, they cannot provide fine-grained spatial semantics.
To address this limitation, we propose a novel Unifying Occupancy (UniOcc) prediction method, explicitly imposing spatial geometry constraint and complementing fine-grained semantic supervision through volume ray rendering.
Our method significantly enhances model performance and demonstrates promising potential in reducing human annotation costs.
Given the laborious nature of annotating 3D occupancy, 
we further introduce a Depth-aware Teacher Student (DTS) framework to enhance prediction accuracy using unlabeled data. Our solution achieves 51.27\% mIoU on the official leaderboard with single model, placing 3rd in this challenge.

\end{abstract}

\section{Introduction}
\label{sec:intro}


CVPR 2023 3D Occupancy Prediction Challenge \cite{occnet} is the world's first benchmark for 3D occupancy in autonomous driving scene perception.
The challenge requires accurate estimation of voxel occupancy and semantics in 3D space using provided surround view images.


In this challenge, we introduce UniOcc, a general solution that utilizes volume rendering to unify 2D and 3D representation supervision, improving multi-camera occupancy prediction models. Rather than designing new model architectures, we focused on enhancing existing models \cite{occnet, bevdet, openoccupancy} in a general and plug-and-play manner.

We elevated the occupancy representation to a NeRF-style representation \cite{plenoxels, dvgo, mipnerf360}, enabling the use of volume rendering to generate 2D semantic and depth maps. This allowed us to perform fine-grained supervision at the level of 2D pixels. The rendered 2D pixel semantic and depth information were obtained by sampling along rays passing through 3D voxels. This explicit integration of geometric occlusion relationships and semantic consistency constraints provides the model with explicit guidance and ensures adherence to these constraints.

It is worth mentioning that UniOcc holds the potential to reduce reliance on expensive 3D semantic annotations. Models trained solely with our volumn rendering supervision, \textbf{without 3D occupancy labels}, even outperformed models trained with 3D label supervision. 
This highlights the exciting potential of reducing the reliance on expensive 3D semantic annotations, as scene representation can be learned directly from affordable 2D segmentation labels.
Furthermore, the cost of 2D segmentation annotation can be further reduced by leveraging advanced technologies like SAM \cite{sam} and \cite{dinov2, visionllm}.

We also introduced the Depth-aware Teacher Student (DTS) framework, a self-supervised training approach. Unlike the classical Mean Teacher \cite{meanteacher}, DTS enhances depth predictions from the teacher model, enabling stable and efficient training while leveraging unlabeled data. Additionally, we applied several simple yet effective techniques to enhance the model's performance. This includes utilizing visible masks during training, employing stronger pretrained backbones, increasing voxel resolution, and implementing Test-Time Augmentation (TTA).

Our solution finally achieves 51.27\% mIoU with a single model and wins 3rd place in this challenge.

\begin{figure*}[ht]
\vspace{-0.4cm}
\centering
\includegraphics[width=1.0\linewidth]{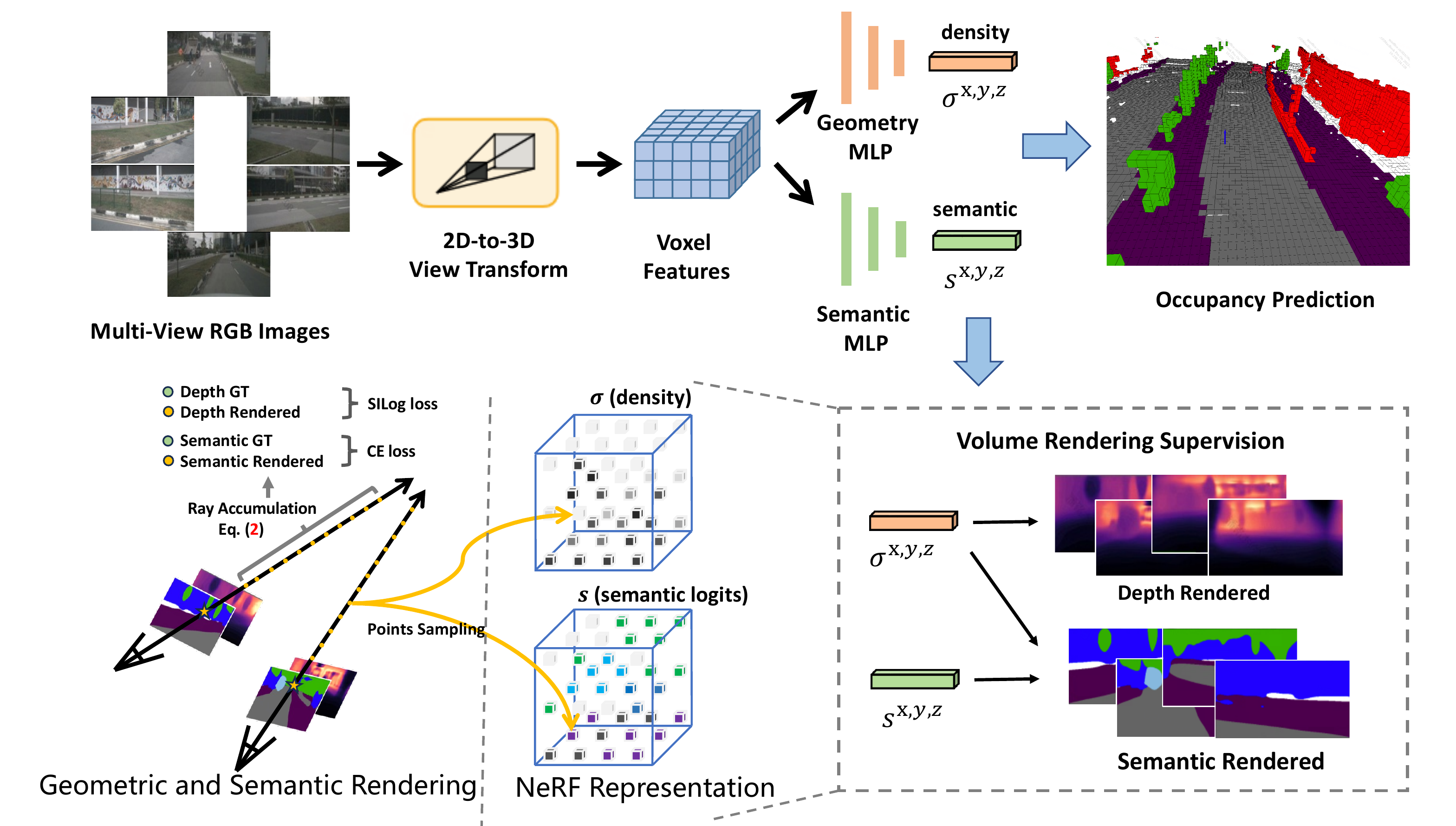}    
\vspace{-0.5cm}
\caption{\label{fig:1}\textbf{An overview of our UniOcc framework.}
}
\label{framework}
\vspace{-0.4cm}
\end{figure*}

\section{Our Solution}

\label{sec:method}

\subsection{Baseline Occupancy Models}
The latest occupancy methods \cite{tpvformer, openoccupancy, voxformer, occ3d, occnet} follow a similar framework as BEV (Bird's Eye View) methods \cite{bevformer, bevdet, bevdepth, petr}, comprising four modules: (1) 2D image feature encoder, (2) 2D-3D view transformer, (3) 3D encoder, and (4) occupancy head.

Since our method is general and not limited to specific model structures, we uniformly represent the first three modules with $F$ and extract 3D voxel features $V^{x,y,z}$ from multi-view RGB images.
\begin{equation}
\begin{aligned}
V = F(I^1,I^2,...,I^n), \\
\end{aligned}
\end{equation}

For occupancy head, existing methods usually use an MLP to classify voxels, predict the logits of $K+1$ categories, and apply cross-entropy loss with semantic occupancy labels for supervision. $K$ represents the number of semantic categories of the dataset, and the last category is the probability of voxel being free.

\subsection{UniOcc}
\subsubsection{From Occupancy to NeRF Representation}
Recently, voxel-based NeRFs methods \cite{plenoxels, dvgo, instant-nerf} have demonstrated powerful performance and the voxel-grid representation is similar to occupancy. 
The goal of this stage is to transfer occupancy representation to NeRF-style representation for supporting volumn rendering. As shown in Fig. \ref{framework}, we first separate the voxel logits predicted by the occupancy head, representing the probability of a voxel being occupied and the corresponding semantic logits. This process is equivalent to using two independent MLPs:
\begin{equation}
\begin{aligned}
\sigma^{x,y,z} = {MLP}_1(V^{x,y,z});\\  s^{x,y,z} = {MLP}_2(V^{x,y,z}), \\
\end{aligned}
\end{equation}
where $\sigma$ denotes density of voxels, and $s$ denotes the semantic logits of occupied voxels. 
\vspace{-1mm}
\subsubsection{Volumn Rendering Supervision}
Based on scene representation $\sigma$ and $s$, we employ geometric and semantic rendering to generate 2D depth and semantic logits. By leveraging camera intrinsics and external parameters, we compute the corresponding 3D ray for each pixel in the 2D image. Subsequently, we sample $N$ 3D points along the ray, denoted as $\{z_k\}^N_{k=1}$. 
Next, we utilize Eq. (\ref{depth_eq}) and Eq. (\ref{semantic_eq}) to obtain rendered depth $D$ and semantic logits $S$ at the pixel level:
\vspace{-1mm}
\begin{align}
\label{eq_ray_rendering}
 T(z_k) &= exp(-\sum_{t=1}^{k-1}\sigma(z_t)\beta_t), \\
\alpha(z_k) &= 1-exp(-\sigma(z_k)\beta_k), \\
 \label{depth_eq}
 D &= \sum_{k=1}^{N}T(z_k)\alpha(z_k)z_k, \\
\label{semantic_eq}
 S &= \sum_{k=1}^{N}T(z_k)\alpha(z_k)s(z_k), 
\end{align}
where $\beta_k=z_{k+1}-z_k$ is the distance between two adjacent sampled points along the ray. 
Considering that autonomous driving is an unbounded and wide-ranging scenario, the effect of using uniform sampling is poor and inefficient. Therefore, we introduced the sampling strategy of \cite{mipnerf360} and achieve voxel-based nonlinear sampling strategy.

For \textbf{geometric rendering}, we utilize the Scale-Invariant Logarithmic (SILog) loss \cite{SLlog} to supervise the rendered depth explicitly, imposing geometric constraints on the model to consider occlusion relationships between voxels.
For \textbf{semantic rendering}, we employ the cross-entropy loss to supervise the rendered semantic logits, uncovering pixel-level fine-grained semantic information.
\vspace{-0.3cm}
\paragraph{Enriching Perspectives with Temporal Frames.}
Leveraging multi-view consistency can greatly enhance the effectiveness of rendering supervision, enabling the model to take into account the occlusion relationships between voxels. However, the range of overlap between surrounding cameras is limited, resulting in a significant portion of voxels being unable to be sampled by multiple rays simultaneously. To address this limitation, we introduce temporal frames as supplementary perspectives. Taking into account the presence of moving objects, we utilize semantic categories to filter out such objects in adjacent frames, thereby only rendering rays for static objects.

\subsection{Depth-aware Teacher Student Framework}
In the data-driven era, data scale and quality play a crucial role in determining model performance. To harness the untapped potential of unlabeled data, we employ semi-supervised learning. However, the traditional approach of MeanTeacher \cite{meanteacher} yielded limited benefits in our experiments. As a result, we developed a new and robust semi-supervised learning framework called Depth-aware Teacher Student (DTS) to address the limitations. As shown in Fig. \ref{fig:1}, DTS builds upon the architecture of Mean Teacher \cite{meanteacher}, where the teacher model guides the student model by generating voxel features and pseudo occupancy labels. The teacher model's parameters are updated using the Exponential Moving Average (EMA) of the student model.

To enhance the teacher model's prediction accuracy, we integrate LiDAR-projected depth ground truth (GT). This provides consistent and effective guidance to the student model during training. Our method involves determining the corresponding index in the LSS depth frustum for the depth label.
We then set the probability of the corresponding position in the predicted depth distribution to 1. While this method may appear intricate, we have observed its remarkable effectiveness. Consistently, it enables the teacher model to surpass the student model by approximately 1 mIoU, thereby ensuring training stability. Similar phenomena have also been discussed in \cite{bevuda,bevdepth}.


\label{sec:formatting}
\begin{figure}[t]
\centering
\vspace{-0.3cm}
\includegraphics[width=1.0\linewidth]{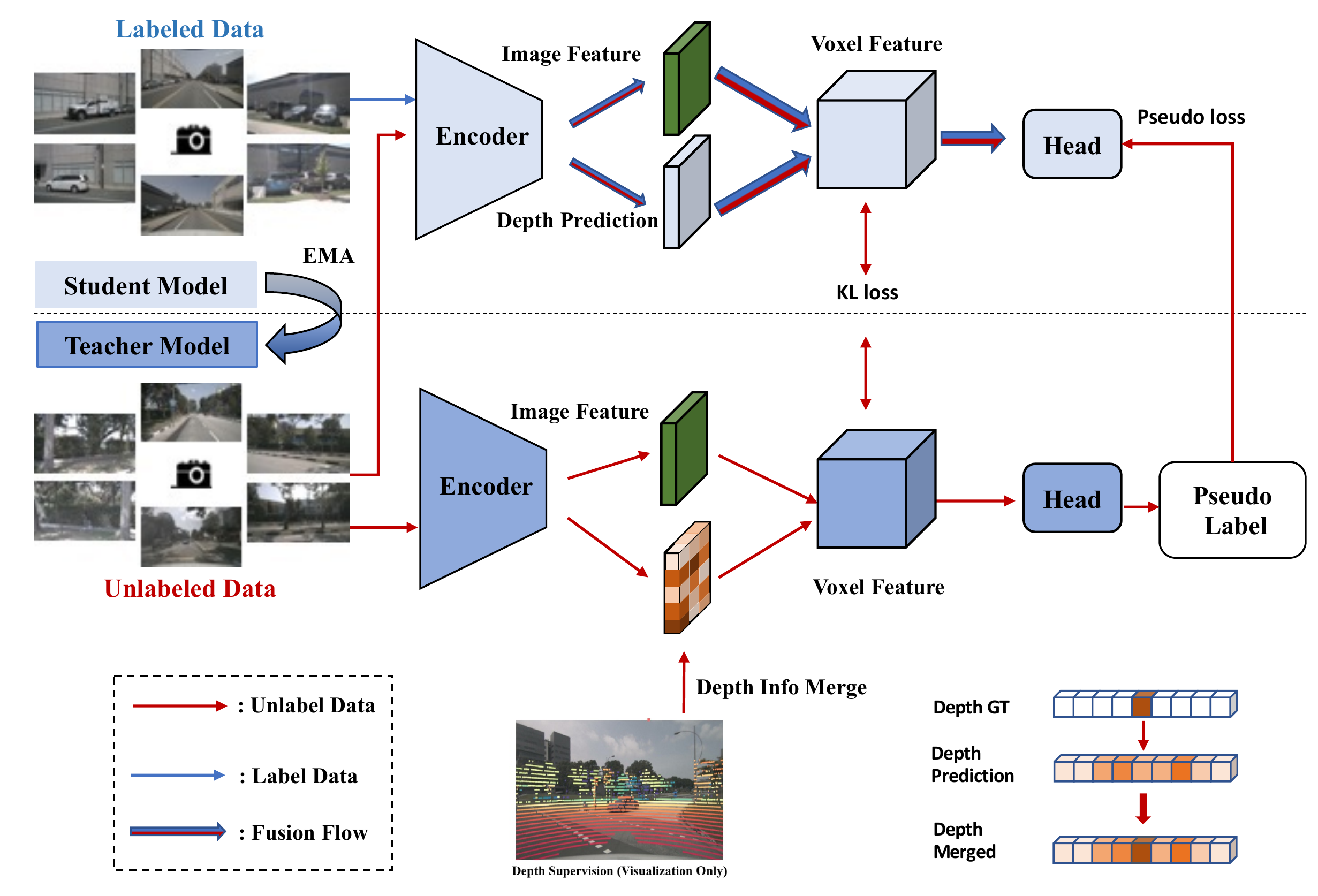}     
\caption{\label{fig:dts}\textbf{Depth-aware Teacher Student framework.} 
}
\vspace{-0.5cm}
\end{figure}


\subsection{Further Improvement}
\paragraph{Usage of visibility masks}
We followed open-source baseline \cite{bevdet} and employed mask\_camera for loss during training. 
The challenge dataset provides visibility masks indicating voxel observations in the current views, determined via ray-casting. Incorporating these masks during training significantly eases the model's learning process. However, this approach has a trade-off, as the model may overlook areas beyond the visible region, like obstacles and the sky, resulting in reduced visualization quality. 
As the competition evaluation only takes into account the visible voxels, the overall evaluation metrics of the model show a substantial improvement.

\vspace{-0.3cm}
\paragraph{Stronger Setting}
We made the following improvements to enhance the performance of the final model. We replaced the 2D backbone with ConvNeXt-Base~\cite{convnext}, utilized stronger pretrain models, increased voxel feature channels, and improved the 3D voxel resolution.



\begin{table*}[!ht]
\vspace{-0.3cm}
\centering
\caption{\textbf{Ablation Study on Validation Set with our improvements.} Rendering Sup (Volumn Rendering Supervision). DTS (Depth-aware Teacher). TTA(Test-Time Augmentation).}
\vspace{-1mm}
\resizebox{2.0\columnwidth}{!}{
\setlength{\tabcolsep}{3.0mm}
\label{main_table}

\renewcommand\arraystretch{1.0}
\begin{tabular}{cccccc|c|c}
\hline
Baseline & Visibility Mask & Rendering Sup & Stronger Setting & DTS & TTA & Resolution & mIoU\\ \hline
\checkmark & & & & & & \multirow{3}{*}{256 $\times$ 704} &19.6  \\
\checkmark & \checkmark & & & & &  &36.1 \\ 
\checkmark & \checkmark & \checkmark &  & & &  &39.7  \\ \hline
\checkmark & \checkmark & \checkmark &  &  & & \multirow{4}{*}{640 $\times$ 1600} &45.2  \\
\checkmark & \checkmark & \checkmark & \checkmark &  & &  &50.4  \\
\checkmark & \checkmark & \checkmark & \checkmark & \checkmark  & &  &51.5 \\ 
\checkmark & \checkmark & \checkmark & \checkmark & \checkmark & \checkmark  & &\textbf{52.1} \\  
 \hline
\end{tabular}
}
\end{table*}

\begin{table*}[!htb]
\vspace{-0.1cm}
\caption{\textbf{Final results on Testing Set.}}
\vspace{-0.1cm}
\begin{adjustbox}{width=1\linewidth,center=\linewidth}
\label{table_test}
\centering
\setlength\tabcolsep{6pt}

\begin{tabular}{l|ccccccccc}
\hline
\multirow{2}{*}{\textbf{mIoU}} & others & barrier & bicycle & bus & car & construction vehicle & motorcycle & pedestrian & traffic cone \\ \cmidrule(){2-10}
&26.94&56.17&39.55&49.40&60.42& 35.51& 44.77& 42.96& 38.45\\ \hline
\multirow{2}{*}{\textbf{51.27}}& trailer & truck & driveable surface & other flat & sidewalk & terrain & manmade & vegetation \\ \cmidrule(){2-10}
& 59.33& 45.90&83.90&53.53&59.45&	56.58&63.82& 54.98   \\ \hline
\end{tabular}
\end{adjustbox}
\vspace{-2mm}
\end{table*}

\section{Experiments}
\label{sec:exp}

\subsection{Dataset and Evaluation metrics}
3D Occupancy Prediction Challenge Dataset contains 700/150/150 sequences for training/validation/testing in the camera-only semantic occupancy prediction track. Regarding to the image data, each frame contains six surround-view images with resolution of $1600 \times 900$ pixels. For semantic occupancy labels, the voxel size is 0.4m, the perception ranges are [-40.0m, 40.0m] for the X-axis, [-40.0m, 40.0m] for the Y-axis and [-1.0m, 5.4m] for the Z-axis. 
The track uses intersection-over-union (IoU) for per-class evaluation and mean IoU (mIoU).


\subsection{Implementation Details}

\paragraph{Baseline Model}
We used the BEVStereo baseline provided by ~\cite{bevdet} and built upon it by incorporating our method.
The baseline model employed ResNet-50 as the backbone. The voxel size extracted by LSS was set to 0.4 meters, and the feature channel was set to 32.
We follow their official training settings as default, including data augmentation (random flip, scale and rotation), training schedule (2$\times$), image resolutions ($256 \times 704$) and others (AdamW optimizer, EMA, 1e-4 learning rate and batch size 32). 
\vspace{-0.3cm}

\paragraph{Rendering Supervision with UniOcc}
For rendering supervision, we project LiDAR points of the key frame onto image and obtain sparse depth label of 2D pixels. While nuScenes does not have 2D semantic label, we used projected lidarseg labels as 2D semantic label.
During training, only pixels those have depth or semantic labels are rendered and used for calculating rendering loss. For each batch, we randomly sample 25, 600 pixels for ray rendering. 
When both 3D label supervision and rendering supervision exist, we set the weight of the rendering loss to 0.1.
\vspace{-0.3cm}
\paragraph{DTS}
When the model's performance is low, employing a semi-supervised learning method may lead to unstable training and poor results. 
To address this, we fine-tune the model trained on labeled data and carefully control the sampling strategy of dataloader to balance the ratio of labeled and unlabeled data (1:1) in each batch. In all experiments using DTS, we finetune two epochs.


\vspace{-0.3cm}
\paragraph{Stronger Setting}
The ConvNeXt-Base pretrain model was provided by \cite{tigbev} pretrained on nuScenes with 3D detection task.
We increased the voxel feature channels from 32 to 256. The voxel-grid edge length was reduced from 0.4m to 0.32m, resulting in an increased voxel shape from 200 $\times$ 200 $\times$ 16 to 250 $\times$ 250 $\times$ 20. Prior to classification, we applied trilinear interpolation to the voxel feature to match the size of the labels.
\vspace{-0.6cm}
\paragraph{Test-Time Augmentation}
In order to further improve the performance, we apply a combination of Test-Time Augmentations (TTAs). Specifically, we use flipping performed in horizontal or vertical at both image-level and BEV-level. 

\subsection{Ablations \& Discussion}

\begin{table}[t!]
\caption{\textbf{Comparable performance without 3D labels.} 
}
\begin{adjustbox}{width=1.0\linewidth,center=\linewidth}
\label{table_wo_3d}
\centering
\setlength\tabcolsep{5pt}
\begin{tabular}{c|c|cc|c}
\hline
Base Model & Resolution & Occ Sup & Rendering Sup  & mIoU\\ \hline
\multirow{2}{*}{BEVFormer} & \multirow{2}{*}{900 $\times$ 1600}& \checkmark & &23.6  \\
 & & & \checkmark &23.3  \\  \hline
\multirow{3}{*}{BEVStereo} & \multirow{3}{*}{256 $\times$ 704}& \checkmark & &19.6  \\
 & & & \checkmark &20.2  \\ 
 & &\checkmark& \checkmark& 21.9  \\    \hline

 \hline
\end{tabular}
\end{adjustbox}
\vspace{-4mm}
\end{table}

\paragraph{Comparable performance without 3D labels.}

 
In Tab. \ref{table_wo_3d}, we first evaluated UniOcc on two open-source baselines \cite{occnet, bevdet}.
Utilizing UniOcc's Rendering Supervision, we achieve competitive results without using 3D occupancy labels. On BEVStereo, our approach achieves 20.2 mIoU, surpassing the model trained with 3D semantic supervision (19.6 mIoU). Combining Rendering Supervision with 3D labels further enhances performance to 21.9 mIoU. This highlights the potential of reducing reliance on costly 3D semantic annotations, learning scene representation directly from affordable 2D segmentation labels.
We can further reduce the cost of 2D segmentation annotation by leveraging cutting-edge technologies such as SAM \cite{sam} and \cite{dinov2, visionllm}.

\vspace{-0.3cm}
\paragraph{Ablations of best single model}
Tab. \ref{main_table} presents the ablation results of our best single model on the Validation split. After using the visibility mask and improve the model performance to 36.1, all subsequent modules can bring significant benefits, ultimately reaching 52.1.

\section{Final Results on Leaderboard}
For our final submission, we utilized all techniques mentioned in Section \ref{sec:method}, \ref{sec:exp} in our model, and train it on both training
and validation splits. We achieved 51.27\% mIoU on test set, ranking the 3rd on nuScenes image-based 3D occupancy prediction challenge Leaderboard in the Autonomous Driving Challenge on CVPR 2023~\cite{occnet}.
The final results are shown in Tab. \ref{table_test}.

{\small
\bibliographystyle{ieee_fullname}
\bibliography{egbib}
}

\end{document}